\documentclass[letterpaper]{article}

\usepackage{aaai20}
\usepackage{times}  
\usepackage{helvet} 
\usepackage{courier}  
\usepackage{multirow}
\usepackage{graphicx}
\usepackage{subfigure}
\usepackage{booktabs} 
\usepackage{amssymb}
\usepackage{amsfonts}
\usepackage{amsmath}
\usepackage{algorithm}
\usepackage{algpseudocode}
\usepackage{enumitem}
\usepackage[dvipsnames]{xcolor}

\newtheorem{definition}{Definition}
\newtheorem{example}{Example}
\newtheorem{proposition}{Proposition}
\newcommand*{\Perm}[2]{{}^{#1}\!P_{#2}}%

\newcommand{\argmin}{\operatornamewithlimits{\textit{arg}\,min}}

\title{One-Shot Induction of Generalized Logical Concepts via Human Guidance}
\author{Mayukh Das\\
University of Texas, Dallas\\
mayukh.das1@utdallas.edu
\And
Nandini Ramanan\\
University of Texas, Dallas\\
nandini.ramanan@utdallas.edu
\And
Janardhan Rao Doppa\\
Washington State University\\
jana.doppa@wsu.edu
\And
Sriraam Natarajan\\
University of Texas, Dallas\\
sriraam.natarajan@utdallas.edu
}
\begin{document}
\maketitle

\author{}

\begin{abstract}
We consider the problem of learning generalized first-order representations of concepts from a single example. To address this challenging problem, we augment an inductive logic programming learner with two novel algorithmic contributions. First, we define a distance measure between candidate concept representations that improves the efficiency of search for target concept and generalization. Second, we leverage richer human inputs in the form of advice to improve the sample-efficiency of learning. We prove that the proposed distance measure is semantically valid and use that to derive a PAC bound. Our experimental analysis on diverse concept learning tasks demonstrates both the effectiveness and efficiency of the proposed approach over a first-order concept learner using only examples. 


\end{abstract}

\section{Introduction}

We consider the problem of learning generalized representations of concepts using a small number of examples. More specifically, we study the case of learning from {\em one example}, which is traditionally called {\em one-shot} learning. This problem has received much attention from ML community~\cite{lake2011,khan2014}. 
We consider a challenging setting inside one-shot learning, that of learning explainable and generalizable (first-order) concepts. These concepts can then be particularly reused for learning compositional concepts, for instance, plans. In our concept learning setting, plan induction becomes a special case where a generalizable plan is induced from a single (noise-free) 
demonstration.
{As an example, consider building a tower that requires learning {\em L-shapes} as a primitive. In our formulation, the goal is to learn a {\em L-shape} from a single 
demonstration. Subsequently, using this concept, 
the agent can learn to build a rectangular base (with 2 {\em L-shapes}) from another single 
demonstration and so on till the tower is fully built.}

Concept learning has been addressed previously in several ways: problem solving by reflection~\cite{stroulia1994learning}, mechanical compositional concepts~\cite{wilson1994geometric}, learning probabilistic programs~\cite{lake2015human}, etc. However, all these methods require a significant number of examples. Concept learning as one-class classification has been considered previously where no negative examples are explicitly created. As one would expect, they are considered while learning an SVM or Nearest Neighbor~\cite{tax2002one}, or with a neural network~\cite{Kozerawski18CVPR}. 

While these methods are closely-related, our work has two key differences. First, we aim to learn an {\em easily interpretable, explainable, and generalizable} concept representation that can then be compounded to learn more complex concepts. To this effect, we focus 
on learning first-order Horn clause~\cite{horn1951sentences}(If Then statements). Second, and perhaps most important, we do not assume the existence of a simulator (for plans) or employ a closed-world assumption to generate negative examples. Inspired by Mitchell's observation of futility of bias-free learning \cite{mitchell1997machine}, we develop an approach that employs domain expertise as inductive bias. The principle of structural risk minimization~\cite{vapnik1999overview} shows how optimal generalization from extremely sparse observations is quite hard. The problem is even more critical in structured domains (most relations are false in the world). Thus, one-shot induction of generalized logical concepts is challenging. We aim to solve this problem via an iterative revision of first-order horn clause theories using a novel scoring metric and guidance from a human. \textit{In essence, we emulate a `student' who learns a generalized concept from an example demonstration provided by the `teacher', by both reflecting as well as, occasionally, asking relevant questions.}

We propose a novel approach referred as {\em Guided One-shot Concept Induction} (\textsc{Goci}) for one-shot concept learning that learns generalized first-order rules. \textsc{Goci} builds upon an inductive logic program (ILP) learner \cite{muggleton1991inductive} and introduces two key algorithmic steps. First, a modified scoring function that allows for explicitly computing distances between concept representations. We show that this distance corresponds to the well known {\em Normalized Compression Distance} (NCD) in the case of plan induction. Based on this observation, we demonstrate that the proposed scoring function is indeed a valid distance metric. Second, the use of domain knowledge from human expert as an inductive bias. Unlike many advice taking systems that employ domain knowledge before training, 
our \textsc{Goci} algorithm identifies the relevant regions of the concept representation space and {\em actively} solicits guidance from the human expert to find the target concept in a sample-efficient manner. Overall, these two algorithmic modifications allow for more effective and efficient learning using \textsc{Goci} that we demonstrate both theoretically and empirically. 



\vspace{1.0ex}

\noindent {\bf Contributions.} We make a few important contributions: 
(1) We derive a new distance-penalized scoring function inside an ILP learner that allows for computing distances between concepts during guided one-shot concept induction. (2) We treat the human-advice as an inductive bias to accelerate learning. Our ILP learner {\em actively} solicits richer information from the human experts than mere labels. (3) We analyze the theoretical properties of \textsc{Goci}. This includes the validity of the distance metric, showing that NCD between plans is a special case of our metric; and deriving a PAC bound based on Kolmogorov complexity. (4) Finally, we demonstrate the exponential gains in both sample efficiency and effectiveness of the algorithm over standard concept learners on diverse concept induction tasks. 



\section{Background and Related Work}
\paragraph{Concept Learning:} Concept learning has previously been studied from several perspectives. Our approach is closely related to Stroulia \& Goel~\shortcite{stroulia1994learning}'s work which proposes an approach for inducing logical problem-solving concepts by reflection. While our scoring metric in \textsc{Goci} is similar to `reflection', its scope is much broader (problem-solving is a special case) and can be used to learn from sparse observations. Wilson~\shortcite{wilson1994geometric} models concepts about continuous geometric shapes by reasoning with two-dimensional bounding boxes. While we use discrete spatial structures as motivating examples, \textsc{Goci} is not limited to discrete spaces. \textsc{Goci} is also related in spirit to probabilistic (bayesian) program induction for learning decomposable visual concepts~\cite{lake2015human} which illustrates how exploiting decomposability is more effective than 
deep learning frameworks. Our approach leverages not only decomposability but implicit relational structure as well, via a logical representation, allowing for generalized concept classes, including plan induction.
Tom Mitchell defines concept learning as \textit{“inferring a boolean-valued function from training examples of its inputs and outputs”}~\cite{mitchell1997machine}, which has inspired its treatment as a standard classification/regression problem as well.\\
\textbf{One/few-Shot Learning:} All approaches described above require a significant number of examples. Our problem setting requires learning from sparse examples (possibly one). Lake et al.,~\shortcite{lake2011} propose a one-shot version of bayesian program induction of visual concepts. There is also substantial work on one/few-shot learning (both deep and shallow) in a traditional classification setting~\cite{bart2005single,vinyals2016matching,wang2018low}, most of which either pre-train with gold-standard support example set or sample synthetic observations under certain assumptions. We make no such assumptions about synthetic examples.\\
\textbf{Theory Induction:} ILP~\cite{muggleton1991inductive,muggleton1994inductive} inductively learns a logical program (first-order theory) that aims to cover most of the positive examples and none of the negative examples. With ILP, goal is to generalize over instances using background knowledge as search bias by building valid hypotheses about unseen examples. In concept learning, generalization is search through space of candidate inductive hypotheses where induction of single theory requires (1) structuring, (2) searching and (3) constraining space of theories~\cite{Lisi2008}. Research in this area has usually focused on one or more of these dimensions. FOIL~\cite{Quinlan1990} is an early non-interactive learner with the disadvantage that it occasionally prunes some uncovered hypotheses. This is alleviated in systems like FOCL by introducing \texttt{language-bias} in form of user-defined constraints~\cite{FOCL}. Top-down relational ILP systems such as PROGOL~\cite{muggleton1995} \& ALEPH~\cite{srinivasan2007aleph} employ inverse entailment to restrict search space. Later with \textit{Interactive ILP}, learner could pose questions and elicit expert advice which allows pruning large parts of search space~\cite{Sammut1986,Rouveirol1992,Rouveirol1990}. To incorporate new incoming information, ILP systems with \textit{theory revision}, incrementally refine and correct the induced theory~\cite{Muggleton1988,Sammut1986}. While \textsc{Goci} is conceptually similar to ALEPH (uses different search strategies and evaluation functions), it additionally acquires domain knowledge by interacting with human expert incrementally. We extend this to one-shot setting.\\
\textbf{Knowledge-Guided Learning:} Background knowledge in ILP is primarily used as search bias. Strong inductive biases in the form of domain knowledge are required to accelerate learning. Mitchell~\shortcite{Mitchell80} proved that biasing learners is necessary to achieve true generalization over new instances. Although earliest form of knowledge injection can be found in explanation-based approaches~\cite{shavlik89ebnn,towell1994knowledge}, our work relates to preference-elicitation framework~\cite{BoutilierEtAl06} which guides learning via human preferences. Augmented learning with domain knowledge as an inductive bias has long been explored across various modeling formalisms, including SVMs, ANNs ~\cite{fung2003knowledge,towell1994knowledge}, probabilistic logic~\cite{odomaaai15}, and planning \cite{DasEtAl18}. Our human-guided \textsc{Goci} learner aims to extend these directions in the context of learning complex concepts (including plans).

\section{Guided One-shot Concept Induction}

We are inspired by a teacher (human) and student (machine) setting in which a {\em small number of demonstrations are used to learn generalized concepts}~\cite{chick2007teaching}. 
Intuitively, the description provided by a human teacher tends to be modular (can have distinct logical partitions), structured (entities and relations between them), and in terms of known concepts. Hence, a vectorized representation of examples is insufficient. We choose a logical representation, specifically a {\em function-free restricted form of first-order logic (FOL)} that models structured spaces faithfully.
\begin{figure}[t]
    \centering
    \includegraphics[width=\columnwidth]{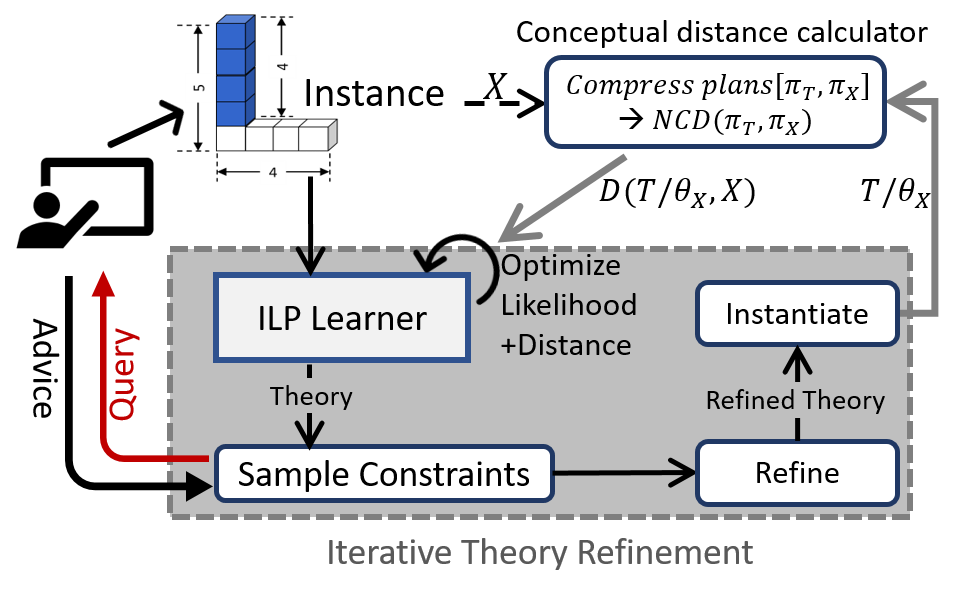}
    \caption{Highlevel overview of our \textsc{Goci} framework.}
    \label{fig:arch}
\end{figure}


Our approach \textsc{Goci} (Figure~\ref{fig:arch}) consists of an ILP learner that induces horn clauses from data. Typical ILP learners use coverage to score candidate clauses.
However, this cannot be used with one-shot learning. Hence, we introduce a penalty function. Specifically, this penalty function is the conceptual distance between the induced theory at the current iteration and the demonstration provided by the ``teacher'' (\textit{Conceptual distance calculator}; top-right in Figure~\ref{fig:arch}). Moreover, the search space of FOL theories is intractable and hence, ILP resorts to employing ``mode" definitions as search bias. However, we need stronger inductive biases in our setting. We allow for a stronger inductive bias by enabling the teacher to provide constraints as advice that can be directly added to the clausal theory. To find the most relevant constraints, \textsc{Goci} queries the expert to choose among a sampled set of constraints (as first-order predicates) and uses the selected constraint to revise the theory. This is analogous to a student asking relevant questions. 

\noindent \fbox{
\parbox{0.97\columnwidth}{
\noindent {\bf Given}: A set of ground predicates (or trajectories) describing one (or few) instance(s) and expert guidance\\
\noindent{\bf To Do:} Induction of first-order logic representation of a concept that generalizes the given example(s) effectively 
}}

\subsection{Input}
\begin{figure}
    \centering
    \includegraphics[width=0.7\columnwidth]{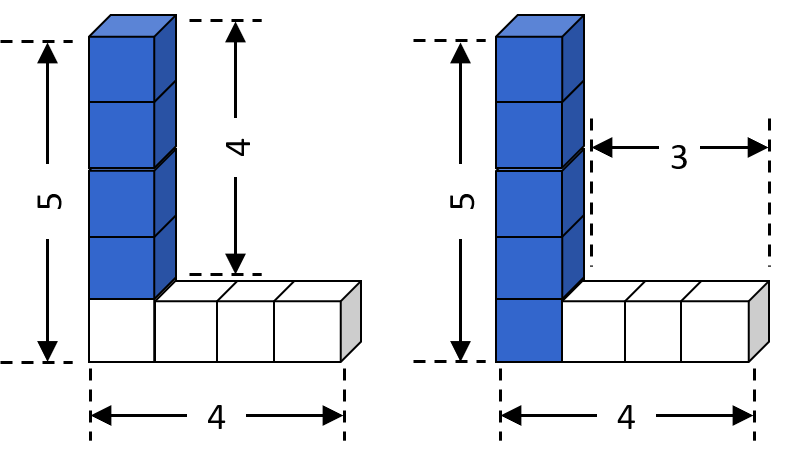}
    \caption{Concept $\mathbb{L}$ ($base=4,height=5$), described as composition of a \textbf{Tower} and a \textbf{Row}}
    \label{fig:Ell}
\end{figure}

The input to \textsc{Goci} is the \textbf{description of the instances(s)} of a concept that the human teacher provides. An input example of a concept is, thus, conjunction of a set of \underline{ground literals (assertions)}. 
\begin{example}
\label{ex:Lex}
An instance of a concept in the domain of spatial structures, can be a structure $\mathbb{L}$ with dimensions $height=5, base=4$ (as shown in Figure~\ref{fig:Ell}).
$\mathbb{L}(S), Height(S,5), Base(S,4),$ $s$ is the concept identifier and may be described as conjunction of ground literals,
\begin{quote}
   $\mathtt{ Row(A)\land}$ $\mathtt{ Tower(B) \land}$ $\mathtt{ Width(A,4) \land}$ $\mathtt{Height(S,5) \land}$ $\mathtt{ Base(S,4) \land }$ $\mathtt{Contains(S,A) \land}$ $Contains(S,B) \land$ $\mathtt{Height(B,4) \land}$ $\mathtt{SpRel(B,A,'NWTop')}$;
\end{quote}
which denotes $\mathbb{L}$ as composition of a `Row' of $w = 4$ and a `Tower' of $h=4$ with appropriate literals describing the scenario (Figure~\ref{fig:Ell} \textbf{left}). As a special case, under partial or total ordering assumptions among the ground literals, an input instance can represent a plan demonstration. 
\end{example}

\subsection{Output}
\textsc{Goci} induces a least general generalization (LGG) horn clause from the \textit{input} example(s). 
\begin{definition}
\label{def:hornconcept} Concept in our setting is represented as horn clause theory.
\begin{align}
    \nonumber T = \mathcal{C}(s^k\ldots):- \bigvee \left[\land_{i=1}^N f_i^k(t_1,\ldots,t_j)\right]
\end{align}
where the body $\land_{i=1}^N f_i(t_1,\ldots,t_j)$ is a conjunction of literals indicating known concepts, the head $\mathcal{C}(s^k\ldots)$ identifies a target concept, and the terms $\{s^k\}$ are logical variables that denote the parameters of the concept. Since a concept can be described in multiple ways (Figure~\ref{fig:Ell}), the final theory will be a disjunction over clause bodies with the same head. A (partial) instantiation of a theory $T$ is $T/\theta$. 
 \end{definition}
Note that these definitions allow for the reuse of concepts induced earlier, potentially in a hierarchical fashion. We believe that this is crucial in achieving human-agent collaboration in complex domains.
\begin{example}
\label{ex:Lconcept}
Figure~\ref{fig:Ell} illustrates an instance of the concept $\mathbb{L}$ which can be described in multiple ways. A possible disjunction could be, 
\begin{align}
    \nonumber \mathbb{L}(s) & :- \mathtt{[Height(s,h_s), Base(s,w_s), Contains(s,a),} \\
    \nonumber & \mathtt{Contains(s,b), Row(a),Tower(b),}\\
    \nonumber & \mathtt{Width(a,w_a),Height(b,h_b), Equal(w_s,w_a),}\\
    \nonumber & \mathtt{Sub(h_b,h_s,1), SpRel(b,a,``NWTop")]}\\
    \nonumber  \bigvee &  \mathtt{[Height(s,h_s), Base(s,w_s), Contains(s,a),} \\
    \nonumber & \mathtt{Contains(s,b), Row(a),Tower(b),}\\
    \nonumber & \mathtt{Width(a,w_{a}),Height(b,h_{b}), Equal(h_{s},h_{b}),}\\
    \nonumber & \mathtt{Sub(w_a,w_{s},1), SpRel(b,s,``W")]}
\end{align}
\end{example}
The generalization must be clearly noted. The last argument of the \texttt{SpRel()} is a constant and not variable, since only this particular spatial alignment is appropriate for the concept of the $\mathbb{L}$ structure. Although the input is a single instance described in one way (Example~\ref{ex:Lex}), \textsc{Goci} should learn a generalized representation such as Example~\ref{ex:Lconcept}. Another interesting aspect are the additional constraints: \texttt{Equal(X,Y)} and \texttt{Sub(X,Y,N)}. While such predicates are a part of the language, they are not typically described directly in the input examples. However, they are essential in appropriate generalization, since they can express complex interactions between numerical (non-numerical) parameters. 

Note that we describe concepts in a generalized manner as horn clause theories. A specific instantiation could be plan induction from sparse demonstrations. In our approach, we can specify the time index as the last argument of both the state and action predicates. Following this definition, we can allow plan induction as shown in our experiments. Our novel conceptual distance is clearer and more intuitive in the case of plans as can be seen later.

\begin{definition}[Decomposable:]
\label{def:decomposable}
A concept $\mathcal{C}$ is Decomposable iff it is expressed as a conjunction of other concepts, and one or more additional literals to model the interactions. $\mathcal{C} \Leftarrow (\bigwedge_i \mathcal{C'}_i) \land (\bigwedge_j B_j)$. Here $\mathcal{C'}_i$ are literals that represent other concepts and $B_j$ are literals that, either describe the attributes of $\mathcal{C'}_i$ or connect them.
\end{definition}

Decomposable means that the concept (currently unknown) can be described as a composition of other known concepts. \textsc{Goci} learns the class of decomposable concepts since it is intuitive for the ``human teacher'' to describe. Decomposable concepts faithfully capture the \textit{modular} and \textit{structured} aspect of how humans would understand and describe instances of concepts in the universe. It also connects to the decomposable nature of planning tasks which can be normally partitioned into sub-tasks that yield sub-plans.
\subsection{Methodology}

\textbf{Search:} Given the definition of the input and output spaces, we now describe the search process. ILP systems perform a greedy search through the space of possible theories. Space is typically defined, declaratively, by a set of mode definitions~\cite{muggleton1995} to guide the search process. Following most well-known ILP systems~\cite{srinivasan2007aleph}, we also start with the most specific clause (known as a bottom clause) from the ground assertions and successively add/modify literals that might improve a rule that best explains the domain. Typically, the best theory is the one which most accurately explains positive example(s) provided while minimizing negative example coverage. Thus, it optimizes the likelihood of a theory $T$ based on the data ($\mathbb{D} = E^+ \cup E^-$). 

We start with a bottom clause, we variablize the ground statements via anti-substitution. Variabilization of theory $T$ is denoted by $\theta^{-1}=\{a/x\}$ where $a\in consts(T)$, $x\notin vars(T)$. That is, inductive substitution $\theta^{-1}$, is a mapping from occurrences of ground terms in theory $T$ to variables.

\vspace{0.5ex}

\textbf{Evaluation Score:} To score a candidate theory, {\em it is necessary to modify the coverage-based ILP scoring} (e.g., ALEPH's compression heuristics) for the following reasons.
\begin{itemize}
\item To use as much of the user-provided advice as possible, we learn long theory, hence the \textit{search space can be exponentially large}. Thus, modes alone aren't enough to guide our search strategy in this situation.
\item There is only one (a few) positive training example(s) to learn from and\textit{ many possible rules can accurately match the training example}. Coverage based scores clearly fail here to identify the best theory. 
\end{itemize}
Most inductive learners optimize some adaptation of likelihood. For a candidate theory $T$, likelihood given data $\mathbb{D}$ is, $LL(T) = \log P(\mathbb{D}|T)$, which is essentially its coverage. In our \textsc{Goci} framework, we have one (at most few) positive example(s). Clearly coverage will not suffice. Hence, we define a modified objective as follows.
\begin{align}
\label{eq:opt}
    T^* = \underset{T \in \tau}{\argmin} \left(-LL(T) + D(T/\theta_X ,  X)\right)
\end{align}
where $T^*$ is the optimal theory, $\tau$ is the set of all candidate theories, and $D$ is the conceptual distance between the \textit{instantiated} candidate theory $T/\theta_X$ and the original example $X$. As explained earlier, a theory $\mathbb{T}$ is a disjunction of horn clauses with the target concept as the head. 

\vspace{0.5ex}

\noindent{\textbf{Distance metric}:} Conceptual distance, $D(T/\theta_X, X)$), is a penalty in our objective. The key idea is that any learned first-order horn clause theory must recover the given instance by equivalent substitution. However, syntactic measures, such as edit distance, are not sufficient since changing even a {\em single literal}, especially, literals that indicate inter-concept relations, could potentially result in a completely different concept. For instance, in blocks-world, the difference between a block being in the middle of a row and one at the end of the row can be encoded by changing one literal. Hence, a more sophisticated semantic distance such as conceptual distance is necessary~\cite{pittphilsci14849}. However, such distances are difficult to compute without a deeper understanding of the domain and its structure. 

Our solution is to employ {\em inter-plan distances}. As discussed earlier, the class of concepts \textsc{Goci} can induce, are decomposable and, hence, are equivalent to parameterized planning tasks. \textbf{One of the key contributions of this work is to exploit this equivalence by using a domain-independent planner to find grounded \textit{plans} for both the theory learned at a particular iteration $i$, $T_i$ and the instance given as input, $X$}. We then compute the Normalized Compression Distance (NCD) between the plans. 

\vspace{0.5ex}

\noindent\textbf{NCD for Plans:} Goldman \& Kuter~\shortcite{goldman2015measuring} proved that NCD is arguably the most robust inter-plan distance metric. NCD is a reasonable approximation of \textit{Normalized Information Distance}, which by itself is not computable \cite{vitanyi2009normalized}. Let the plans for $T_i/\theta_X$ and $X$ be $\pi_T$ and $\pi_X$. To obtain NCD, we execute string compression (lossy or lossless) on each of the plans as well as the concatenation of the two plans to recover the compressed strings $C_T, C_X$, and $C_{T, X}$ respectively. NCD between the plans can be computed as,
\begin{equation}
\label{eq:ncd}
    NCD(\pi_T , \pi_X) = \frac{C_{T,X} - \min(C_T, C_X)}{\max(C_T, C_X)}
\end{equation}
The conceptual distance between a theory $T$ and $X$ is the NCD between the respective plans, $D(T/\theta_X, X) = NCD(\pi_T , \pi_X)$. This entire computation is performed by the \textit{Conceptual distance calculator} as shown in Figure~\ref{fig:arch}.

\vspace{0.5ex}

\textbf{Observations about NCD:} \textbf{(1)} Conceptual distance as a penalty term for negative log-likelihood in the scoring function ensures that the learned theory will correctly recover the given example/demonstration. \textbf{(2)} $D(T/\theta_X, X)$ generalizes to the \underline{\textit{Kolmogorov-Smirnov} statistic} between two target distributions if we induce probabilistic logic theories. We prove these insights theoretically in the next section.

\textbf{Human Guidance:} As outlined earlier, the search space in ILP is provably infinite. Typically language-bias (modes) and model assumptions (closed world) are used to prune the search space. However, it is still intractable with one (or few) examples. Hence, we employ human expert guidance as constraints that can be directly used to refine an induced theory, acting as a strong inductive bias. Also, we are learning decomposable concepts (see Definition~\ref{def:decomposable}). This allows us to exploit another interesting property. Constraints can now be applied over the attributes of the known concepts that compose the target concept, or over the relations between them. Thus, \textsc{Goci} directly includes such constraints in the clauses as literals (see Example~\ref{ex:Lconcept}). Though such constraint literals come from the pre-declared language, they are not directly observed in the input example(s). So an ILP learner will fail to include such literals in the induced clauses. 

If the human inputs (constraints) are provided upfront before learning commences, it turns out to be wasteful/irrelevant. 
More importantly, it places the burden on the human consider all possible scenarios where advice could be needed. 
To alleviate this, {\em our framework explicitly queries for human advice on the relevant constraint literals, which are most useful}. Let $\mathbb{U}$ be a predefined library of constraint predicates in the language, and let $\mathcal{U}() \in \mathbb{U}$ be a relevant constraint literal. Human advice $\mathcal{A}$ can essentially be viewed as a preference over the set of relevant constraints $\lbrace\mathcal{U}()\rbrace$. If $U_\mathcal{A}$ denotes the preferred set of constraints, then we denote the theory having a preferred constraint literal in the body of a clause as $\tau_\mathcal{A}$. \textit{(For instance, based on Example~\ref{ex:Lconcept} \textsc{Goci} could query which of the two sampled constraints $\mathtt{Sub(h_b,h_s,1)}$ \& $\mathtt{Greater(h_b,h_s)}$ is more useful. Human could prefer $\mathtt{Sub(h_b,h_s,1)}$, since it clearly subsumes the other.)} Scoring function now becomes:
\begin{align}
\label{eq:obj}
    \nonumber T^* = \underset{T \in \tau}{\argmin} \left(-LL(T) + D(T/\theta_X , X)\right)~~
   \text{: } \tau \subseteq \lbrace\tau_\mathcal{A}\rbrace
\end{align}
Thus, we are optimizing the constrained form of the same objective as Equation~\ref{eq:opt} which aims to prune the search space. This is inspired by advice elicitation approaches~\cite{odomaaai15}. While our framework can incorporate different forms of advice, we focus on preference over constraints on the logical variables, in our one-shot setting. 
The formal algorithm, described next, illustrates how we achieve this via an iterative greedy refinement (Figure~\ref{fig:arch}, query-advice loop shown in left part of the image). 


\subsection{The \textsc{Goci} Algorithm}

Algorithm~\ref{alg:goci} outlines the \textsc{Goci} framework. It initializes a theory $T_0$, by variablizing the \textit{`bottom clause'} obtained from $X$ and background knowledge [\texttt{lines 3 \& 5}]. Then it performs a standard ILP search (described earlier) to propose a candidate theory [\texttt{line 6}]. This is followed by the guided refinement steps, where constraint literals are sampled (parameter tying guides the sampling) and the \textit{human teacher} is queried for preference over them, such that the candidate theory can be modified using preferred constraints [\texttt{lines 7-9}]. The function \textsc{Ncd}() performs the computation of the conceptual distance, by first grounding the current modified candidate theory $T'$ with the same parameter values as the input example $X$, then generating grounded plans and finally calculating the normalized compression distance between the plan strings (as shown in Figure~\ref{fig:arch} and Equation~\ref{eq:ncd}) [\texttt{line 10}]. Distance-penalized negative log-likelihood score is estimated and minimized to find the best theory at the current iteration [\texttt{lines 11-14}], which is then used as the initial model in the next iteration. This process is repeated either until convergence (no change in induced theory) or maximum iteration bound ($L$).
\begin{algorithm}[t]
\caption{Guided One-shot Concept Induction}
\label{alg:ILR}
\begin{algorithmic}[1]
\Procedure{\textsc{Goci}}{Instance $X$}
\State \textbf{Initialize:} Set Iteration $\ell \gets 1$ 
\State \textbf{Initialize:} Bootstrap theory $T_{0} \gets X/\theta^{-1}$
\Repeat
\State Use $T_{\ell-1}$ as initial model
\State Candidate theory $T_\ell \gets$ \Call{Search}{$T\in\tau | T_{\ell-1}$}
\State \textbf{Sample} applicable constraints $\mathcal{U} \in \mathbb{U}$
\State $\mathcal{U}_\mathcal{A} \gets$ \textsc{Query}($human, \mathcal{U}$)
\State ${T}' \gets T_\ell \oplus \mathcal{U}_\mathcal{A}$ \Comment{$\forall~ \mathcal{U}_\mathcal{A} \in \mathcal{A}$}
\State $D_\ell({T}'/{\theta_X}, X) \gets \Call{Ncd}{\pi_{T'/\theta_X},\pi_X}$
\State Score $\mathcal{S}_\ell \gets \left(-LL(T')+D(T'/{\theta_X},X)\right)$
\If{$S_\ell < S_{\ell -1}$} \Comment{minimize}
    \State Retain ${T}'$: Update $T_{\ell} = {T}'$
\EndIf
\Until{$\ell \leq L$ OR ${T}_{\ell} = {T}_{\ell-1}$}
\EndProcedure
\end{algorithmic}
\label{alg:goci}
\end{algorithm}


\subsection{Theoretical analysis}
\label{sec:theo}
\subsubsection{Validity of Distance Metric: }
Normalized information distance $\delta (x,y)$ between 2 strings $x$ and $y$ is provably a valid distance metric~\cite{vitanyi2009normalized}
$$\delta (x,y) = \frac{\max K(x|y), K(y|x)}{\max K(x), K(y)}$$
 where $K(x)$ is the Kolmogorov complexity of a string $x$ and $K(x|y)$ is the conditional Kolmogorov complexity of $x$ given another string $y$. NCD is a computable approximation of the same [$D(x,y) \approx \delta(x,y)$]. Thus, we just verify if $\delta$ is a correct conceptual distance measure.

Consider two theories $T_Y$ and $T_Z$, with same parameterizations (\textit{i.e.}, same heads). Let $T_Y/\theta$ and $T_Z/\theta$ be their grounded theories with identical parameter values $\theta$. Our learned theories are equivalent to planning tasks. Assuming access to a planner $\Pi()$ which returns $Y= \Pi(T_Y/\theta)$ and $Z=\Pi(T_Z/\theta)$, the two plan strings w.r.t the instantiations of concepts represented by $T_Y$ and $T_Z$ respectively.

\begin{proposition}[Valid Conceptual Distance]
Normalized information distance $\delta(Y,Z)$ is a valid and sound conceptual distance measure between $T_Y$ and $T_Z$, i.e., $\delta(Y,Z) = 0$ iff the concepts represented by $T_Y$ and $T_Z$ are equivalent.
$$(\delta(Y,Z)=0) \iff (T_Y \equiv T_Z)$$
\end{proposition}

\begin{proposition}[Generalization to Kolmogorov-Smirnov]
In generalized probabilistic logic, following Vit{\'a}nyi~\shortcite{vitanyi2013conditional}, $\delta (Y,Z)$ corresponds to two-sample Kolmogorov-Smirnov statistic between two random variables $T_Y/\theta$ and $T_Z/\theta$ with distributions $P_{T_Y}$ and $P_{T_Z}$ respectively. 
$v(T_Y, T_Z)=\textbf{sup}_{\theta \in \mathcal{F}}\left|F_{T_Y}(\theta)-F_{T_Z}(\theta)\right|$,
where $F_{T_Y}()$ is the cumulative distribution function for $P_{T_Y}$ and $\textbf{sup}_{\theta \in \mathcal{F}}$ is the supremum operator. In deterministic setting, $\delta$ is a special case of $v$, 
$\delta (Y,Z) \preceq v(F_{T_Y}, F_{T_Z})$.
\end{proposition}

\subsubsection{PAC Learnability:}
To analyze the theoretical properties of \textsc{Goci}, we build on the PAC analysis for recursive \textit{rlgg} (relative least general generalization) in GOLEM for function-free \textit{horn} clause induction due to Muggleton \& Feng~\shortcite{muggleton1990efficient}. 
Let $n$ denote the sample size and $\mathcal{H}$ denote the hypothesis space. 
{As shown earlier, \textsc{Goci} learns the final theory by iterative refinement after inducing an initial theory}. Denote the initial hypothesis space as $\mathcal{H}_0$ and hypothesis space of the final theory as $\mathcal{H}^*$ (such that $T^* \in \mathcal{H}^*$). 

\vspace{0.5ex}

\begin{proposition}[Sample complexity]
Following Valiant \shortcite{valiant1984theory} and Mooney~\shortcite{mooney1994preliminary}, with probability $(1-\delta)$, the sample complexity of inducing the optimal theory $T^*$ is:
\begin{equation}
 \label{eq:refine}  n^* = \mathcal{O} \left( \frac{1}{\epsilon} \left[d^L \ln(\left|\mathcal{H}_0\right| + d + m) + \ln(\frac{1}{\delta})\right] \right)
\end{equation}
where $\epsilon$ is the regret, $n^*$ - sample complexity of $\mathcal{H}^*$, $m$ - number of distinct predicates, $d$ is the distance of the current revision from the last known consistent theory, and $L$ is the upper bound on the number of refinement steps (iterations). 
\end{proposition}
ILP (such as GOLEM~\cite{muggleton1990efficient}) induces $ij$-Determinate clauses~\cite{muggleton1990efficient}, where $i$ is the maximum depth of a variable in a clause and \& $j$ is the maximum arity. Thus, in our problem setting, 
$ \left| \mathcal{H}_0 \right| = \mathcal{O}( (tpm)^{j^i})$, where $t$ is the number of terms, and $p$ is the place ~\cite{muggleton1990efficient}.
Hence, (if $j$ \& $i$ is bounded : $j^i =c$) Equation~\ref{eq:refine} can be reformulated as: 
\begin{equation}
 \label{eq:refine2} n^* = \mathcal{O} \left( \frac{1}{\epsilon} \left[d^L \ln \left(\left( (tfm)^c\right) + d + m\right) + \ln(\frac{1}{\delta})\right] \right)
\end{equation}

Mooney \shortcite{mooney1994preliminary} defines distance $d$ to be the number of single literal changes in a single refinement step. 
In Algorithm~\ref{alg:goci}, we observe that at each iteration $\ell \leq L$, updates are w.r.t. the preferred constraint predicates $\mathcal{U}_{\mathcal{A}}\in \mathbb{U}$.

\vspace{0.5ex}

\begin{proposition}[Refinement distance]
$d$ is upper bounded by the expected number of literals that can be constructed out of the library of constraint predicates with human advice $\mathbb{E}_{\sim\mathcal{A}}\left[\left|\mathbb{U}\right|\right]$ and lower bounded by the conceptual distance between theory learned at two consecutive iterations since we adopt a greedy approach. If $Pr_\mathcal{A}(\mathcal{U})$ denotes the probability of a constraint predicate being preferred then,
$\left|D_\ell - D_{\ell-1}\right| \leq d \leq \sum_{i=1}^ {2^{(\left|\mathbb{U}\right|-1)} \times {\Perm{t}{q}}} Pr_\mathcal{A}(\mathcal{U}_i)$
where $2^{(\left|\mathbb{U}\right|-1)} \times {\Perm{t}{q}}$ is the maximum possible number of constraint literals and $q$ is the maximum arity of the constraints. If we use only pairwise constraints, then $q$ =$2$.
\end{proposition}

Our input is sparse (one or few instances). \textsc{Goci} elicits advice over constraints to acquire additional information. Let $\left|X\right|$ be the number of input examples. 

\begin{proposition}[Advice complexity]
From Equations \ref{eq:refine} and \ref{eq:refine2}, at convergence $\ell = L$, we get
$\frac{n^* - \left|X\right|}{L}$ examples, on an average, for a concept $\mathcal{C}$ to be PAC learnable using \textsc{Goci}. 
\end{proposition}


\section{Evaluation}
We next aim to answer the following questions explicitly:
\begin{enumerate}[leftmargin=2.2\parindent]
    \item[\textbf{(Q1)}] Is \textsc{Goci} effective in one-shot concept induction?
    \item[\textbf{(Q2)}] How sample efficient is \textsc{Goci} compared to baselines?
    \item[\textbf{(Q3)}] What is the relative contribution of the novel scoring function vs. human guidance towards performance?
\end{enumerate}
We developed our framework by extending Wisconsin Inductive Logic Learner~\cite{natarajanilp}. We modified the scoring function with NCD penalty and integrated a customized SHOP2 planner~\cite{nau2003shop2} for inter-plan NCD computation. We added constraint sampling and human guidance in iterative fashion as outlined in Algorithm~\ref{alg:goci}.
 
\begin{figure}
    \centering
    \subfigure{
    \includegraphics[width=0.47\columnwidth]{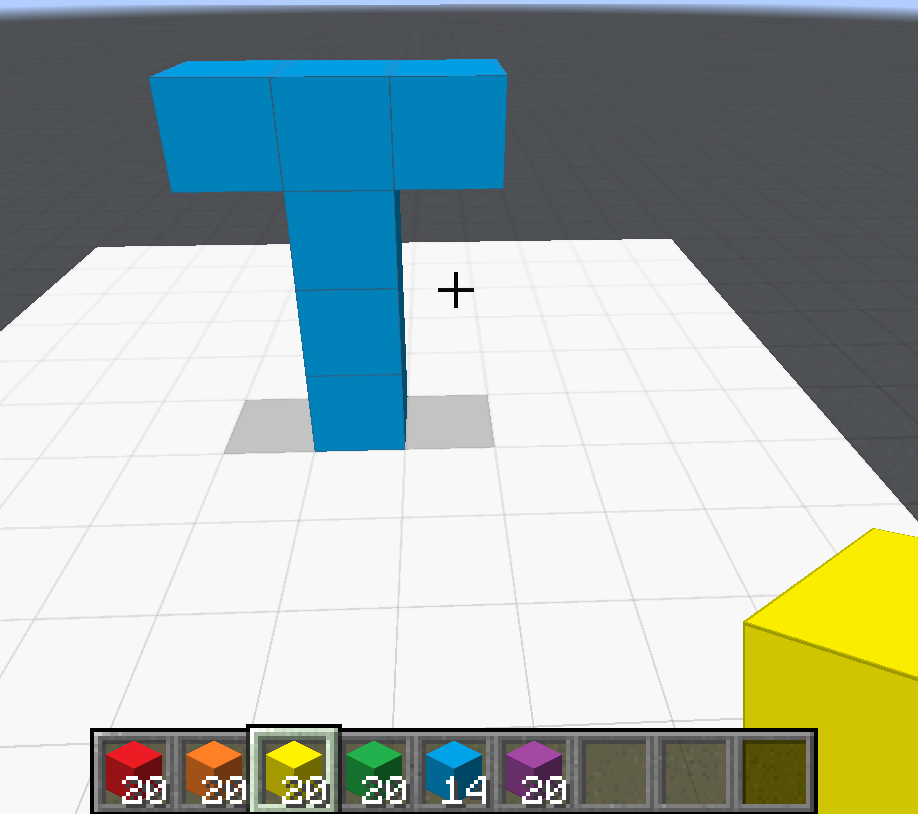}
    }
    \subfigure{
    \includegraphics[width=0.47\columnwidth]{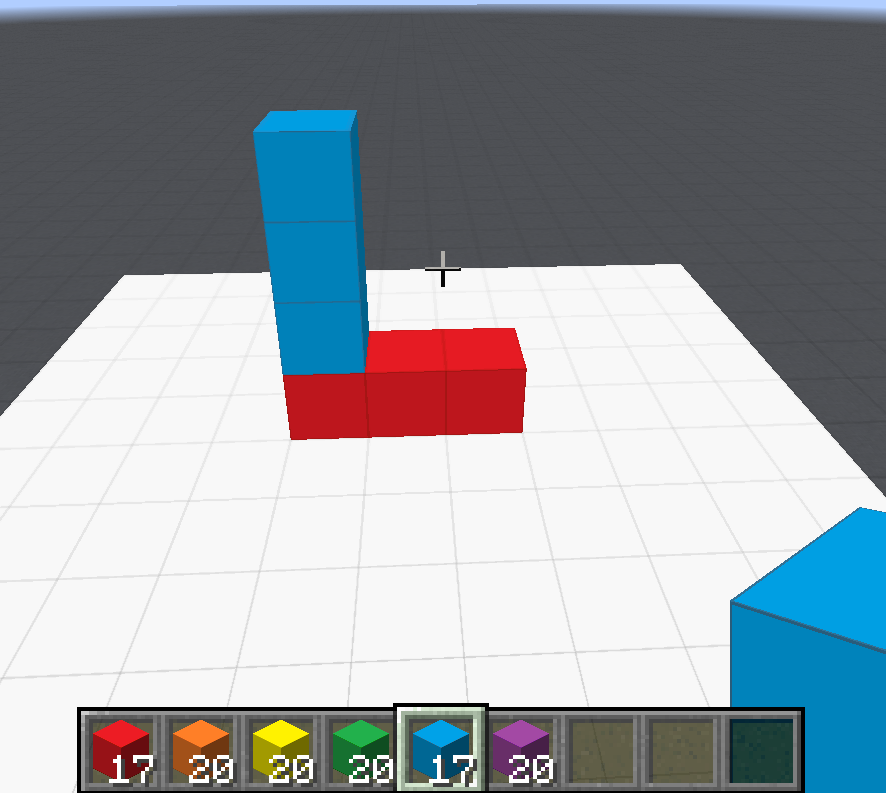}
    }
    \caption{Instances of spatial concepts in Minecraft. \textit{(Left)} Upright Tee, \textit{(Right)} Upright L}
    \label{fig:minecraft}
\end{figure}

\begin{figure*}[h]
    \centering
    \subfigure[Minecraft domain]{
    \includegraphics[width=0.32\textwidth]{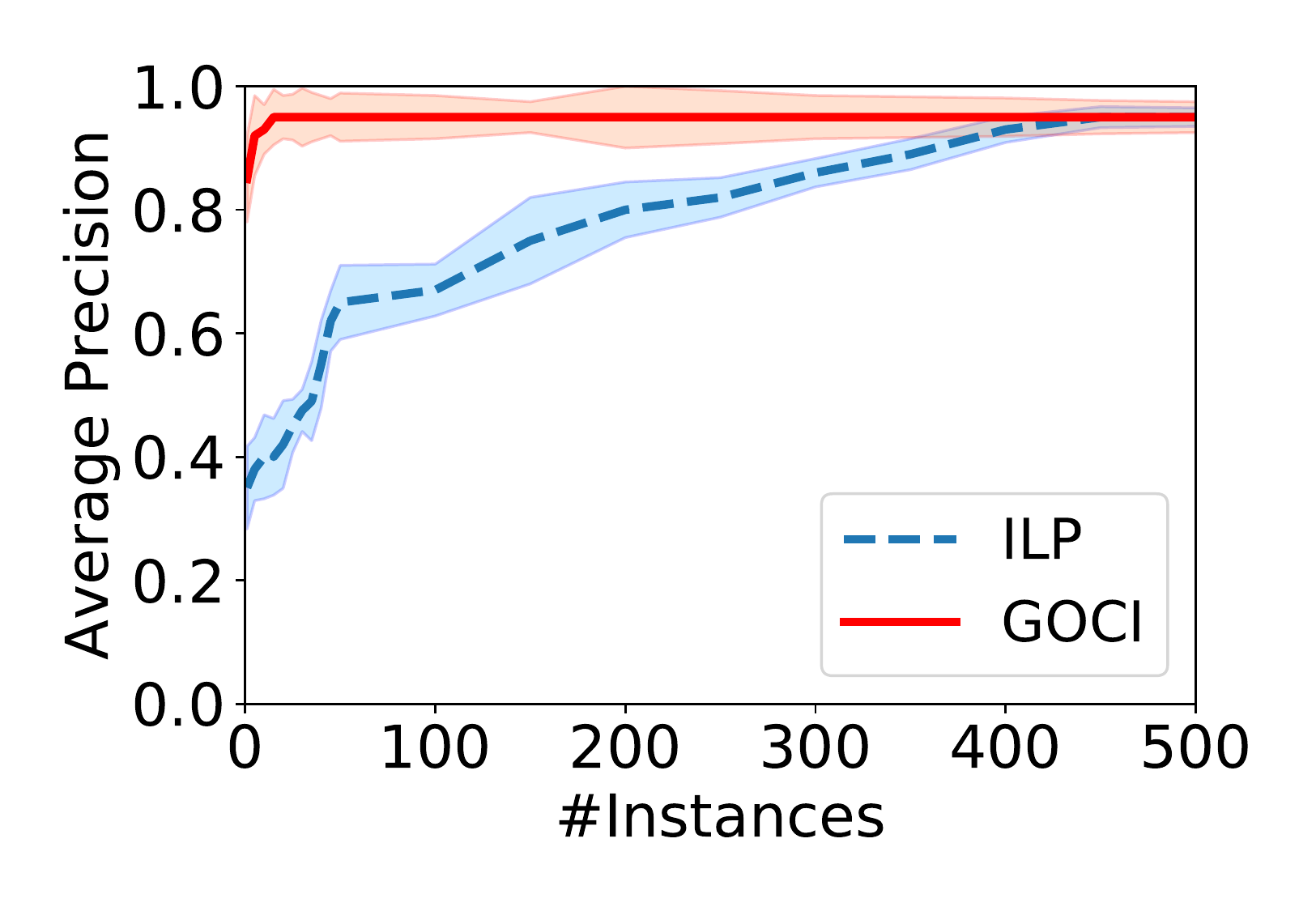}
    \label{fig:LCBlocks}}
    \subfigure[Assembly domain]{
    \includegraphics[width=0.32\textwidth]{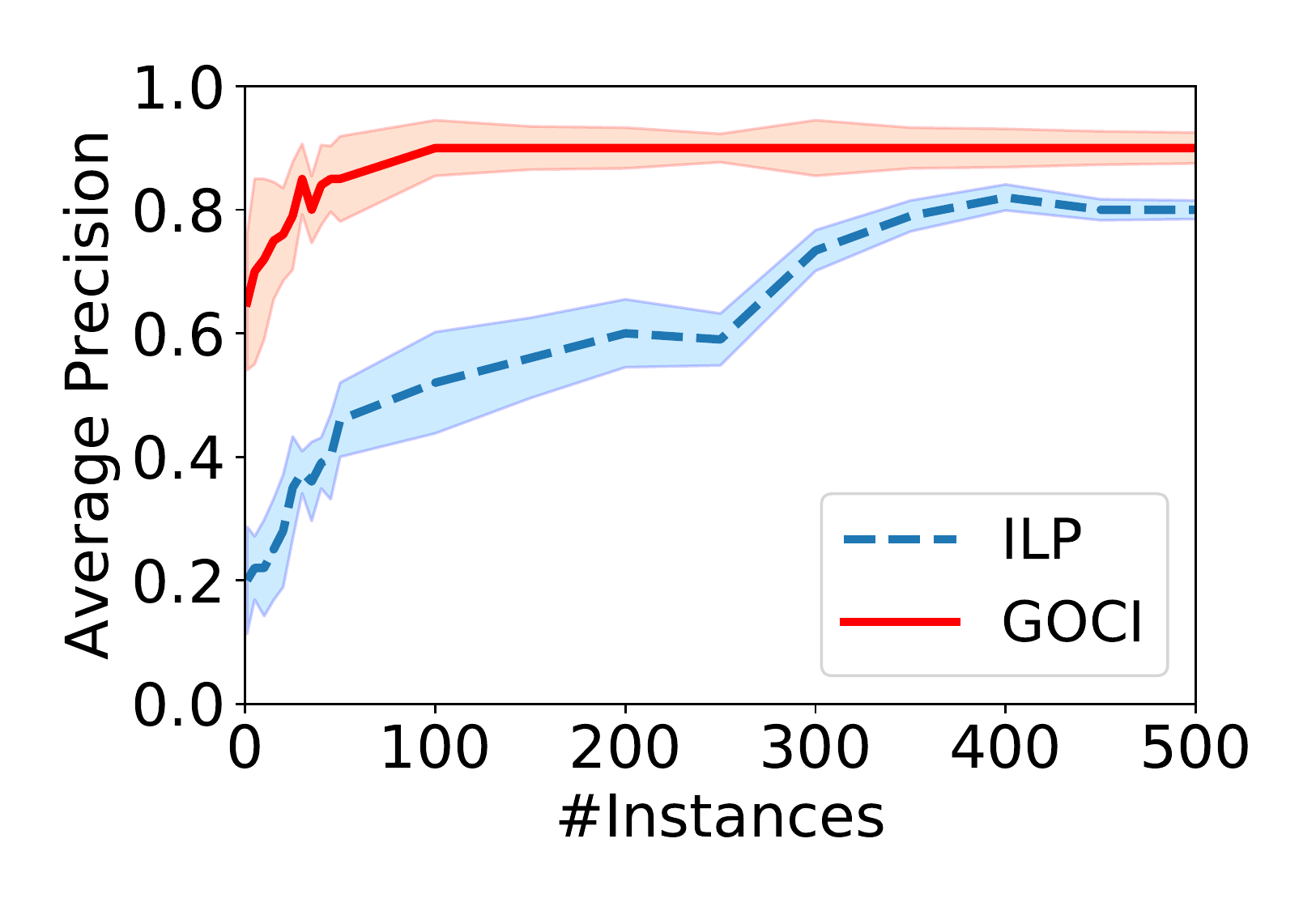}
    \label{fig:LCAssembly}}
    \subfigure[ChEBI]{
    \includegraphics[width=0.32\textwidth]{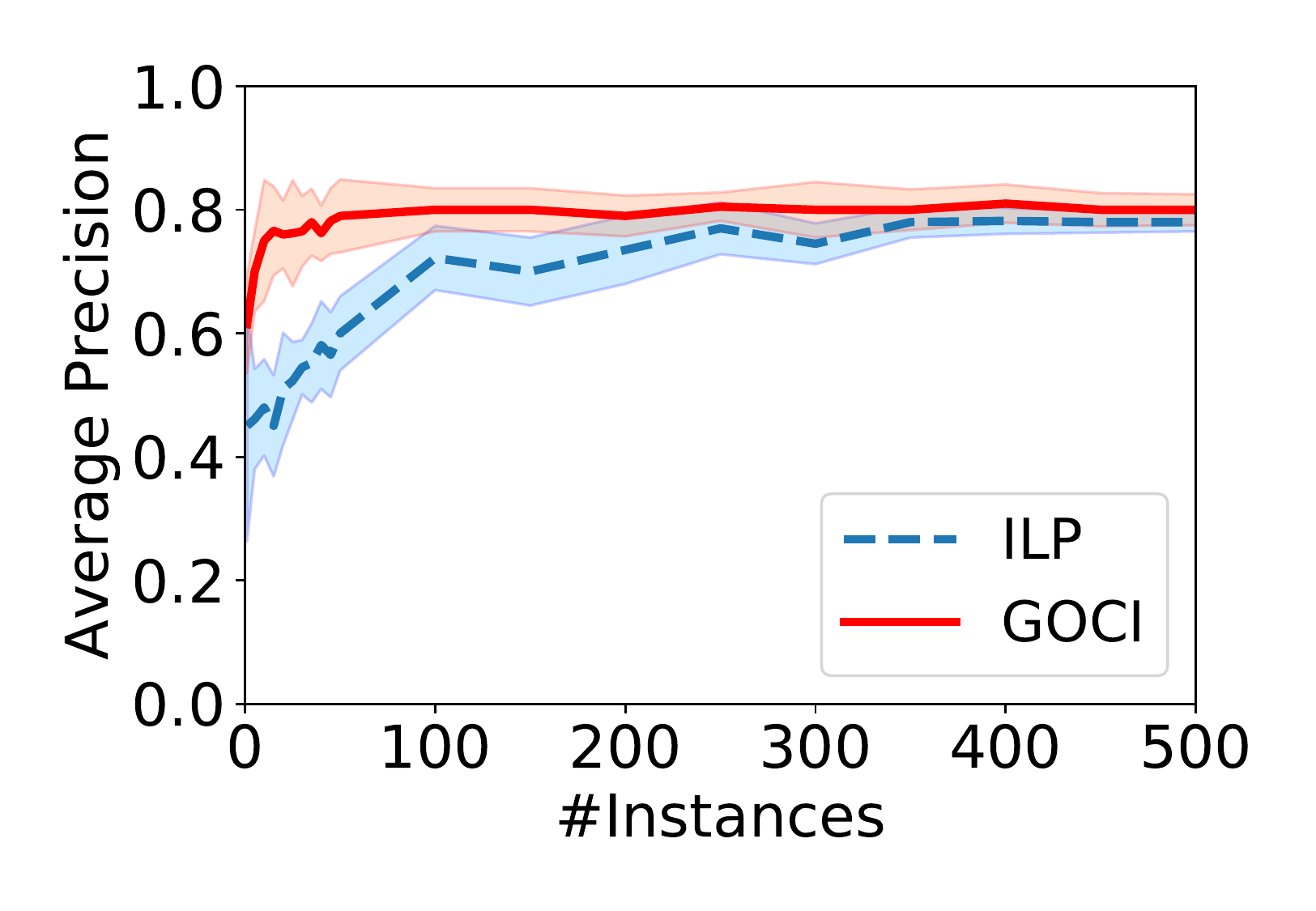}
    \label{fig:LCchebi}
    }
    \caption{Learning curves for varying sample size to compare the sample-efficiency of \textsc{Goci} and ILP (best viewed in color).}
    \label{fig:LC}
\end{figure*}

\begin{figure}[t]
    \centering
    \includegraphics[width=0.8\columnwidth]{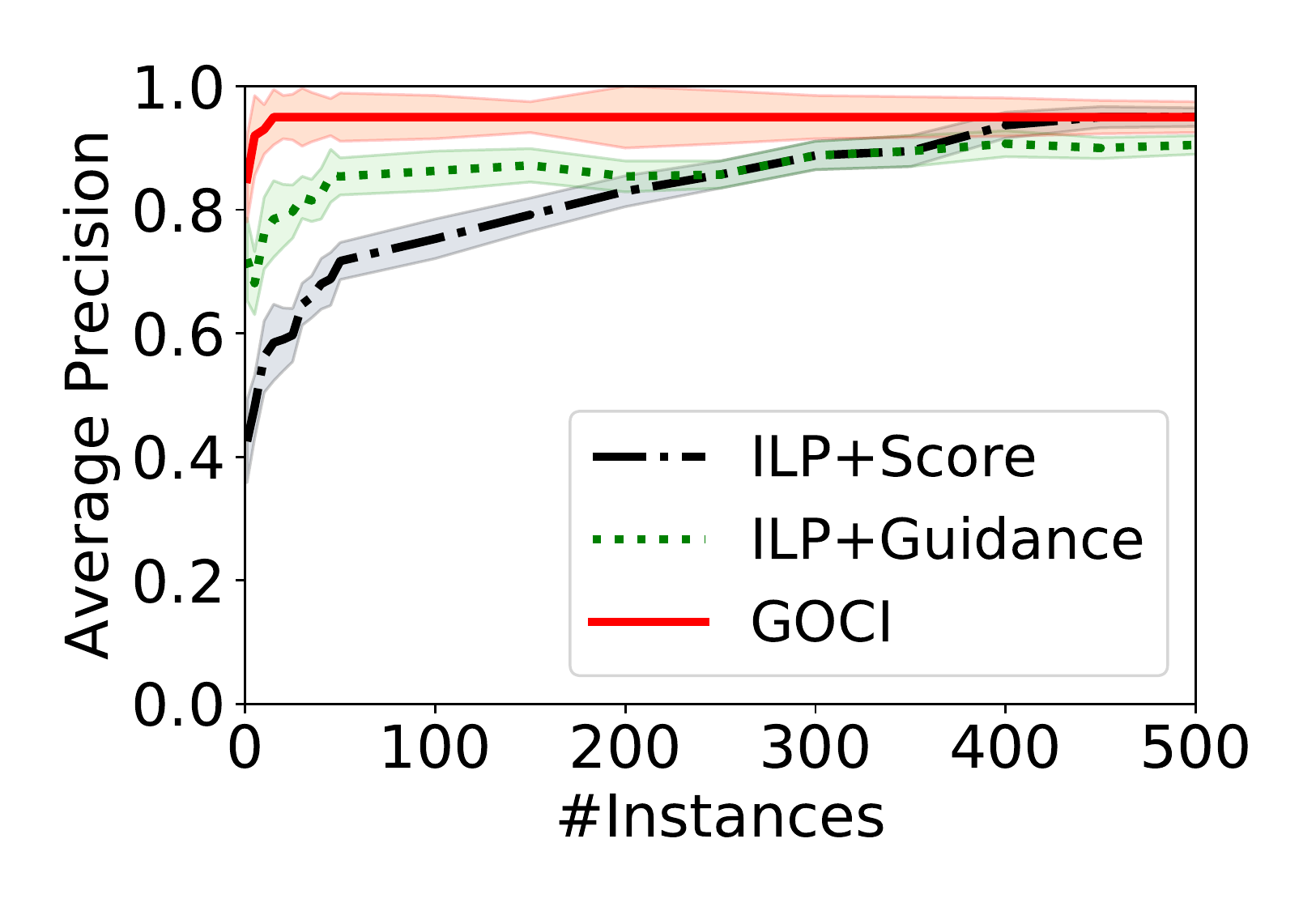}
    \caption{Results of ablation study on Minecraft domain. Relative contribution of our distance-penalized score vs. human guidance (best viewed in color).}
    \label{fig:componentCompare}
\end{figure}

\vspace{0.5ex}

\noindent\textbf{Experimental Setup:} We compare \textsc{Goci} with a standard ILP system with no enhancements. 
We focus on the specific task of ``one-shot concept induction'', with a single input example for each of the several types of concepts and report aggregated precision. We consider precision because preference queries are meant to eliminate false positives in our case. To demonstrate the robustness of \textsc{Goci} w.r.t sample complexity, beyond one-shot case, we conducted experiments with varying sample sizes for each concept type across all domains and show learning curves for the same. We also evaluate the relative contribution of two important components of our proposed framework: (a) novel scoring metric and (b) human guidance; \textit{i.e.}, we compare against two more baselines (\textit{ILP+Score} and \textit{ILP+Guidance}). For every domain, we consider ten different types of concepts (ten targets). All the results are aggregated over five different runs.

\vspace{0.5ex}

\noindent\textbf{Domains} We employ 3 domains with varying complexity:

\vspace{0.4ex}

{\bf \textit{1. Minecraft (Spatial Structures):} }The goal is to learn discrete spatial concepts in a customized~\cite{minedialog} Project Malmo platform for Minecraft. 
Dialogue data in Malmo is available online, and we converted them into a logical representation. All structures are in terms of discrete atomic unit blocks (cubes). Figure~\ref{fig:minecraft} shows examples of some simple spatial structures that \textsc{Goci} was able to learn. The representation language is similar to the Example~\ref{ex:Lconcept}, with some pre-existing concepts in knowledge base such as \texttt{Row}, \texttt{Tower}, \texttt{Cube} etc.

\vspace{0.4ex}

{\bf \textit{2. Assembly (a planning domain):}} Generalized concepts are equivalent to parameterized planning tasks. Assembly is a well-known planning domain, where different mechanical structures (machines) are constructed using different parts and resources. Input is a conjunction of ground literals indicating ground plan demonstration (assuming total ordering). 

\vspace{0.4ex}

{\bf \textit{3. Chemical Entities of Biological Interest (ChEBI): }}
ChEBI~\cite{degtyarenko2007chebi} is a compound database which contains some important structural features and activity-based information for classification of chemicals. Core information ChEBI provides are (1) Molecular structure, (2) Biological role, (3) Application, and (4) Subatomic particle, which may be suitable for the prediction of various attributes of compounds. We model the Benzene molecule prediction task following the description in ChEBI. The data has predicates such as \texttt{Carbon}, \texttt{SingleBond}, \texttt{DoubleBond}, \texttt{HasAtom} etc.

\begin{table}
    \centering
    \begin{tabular}{|l|c|c|c|}
    \hline
        \textbf{Domain} & \textbf{Approach} & \textbf{Avg. Precision} & \textbf{\#Queries} \\
        \hline
        \hline
         \multirow{ 2}{*}{Minecraft} & \textsc{Goci} & $0.85$ & $5.5 \pm 3$\\
         & ILP & $0.35$ & - \\
         \hline
         \multirow{ 2}{*}{Assembly} & \textsc{Goci} & $0.65$ & $16.5\pm 4$\\
         & ILP & $0.2$ & - \\
         \hline
         \multirow{ 2}{*}{ChEBI} & \textsc{Goci} & $0.615$ & $13.1\pm 2.13$\\
         & ILP & $0.45$ & - \\
         \hline
    \end{tabular}
    \caption{Results for one-shot concept learning.}
    \label{tab:1shotresult}
\end{table}

\subsection{Experimental Results}

\noindent\textbf{[Effective One-shot (Q1)]} Table~\ref{tab:1shotresult} shows the performance of \textsc{Goci} on one-shot concept learning tasks as compared to standard ILP. \textsc{Goci} significantly outperforms ILP across all domains
answering (Q1) affirmatively. 
Also, note that \textsc{Goci} is very `query' efficient as observed from the small average number of queries posed in the case of each domain. 

\vspace{0.1ex}

\noindent\textbf{[Sample Efficiency (Q2)]} In figure~\ref{fig:LC}, we observe that \textsc{Goci} converges within significantly smaller sample size across all domains, thus, supporting our theoretical claims in Section~\ref{sec:theo}. In ChEBI, though, quality of planning domain encoding might explain mildly lower-precision yet \textsc{Goci} does perform significantly better than vanilla ILP learner. 

\vspace{0.1ex}

\noindent\textbf{[Relative contribution (Q3)]} Figure~\ref{fig:componentCompare} validates our intuition that both components (scoring function and human-guidance) together make \textsc{Goci} a robust one-shot (sample-efficient) concept induction framework. Though human guidance, alone, is able to enhance the performance of a vanilla ILP learner in sparse samples, yet it is not sufficient for optimal performance. In contrast, although the advantage of our novel distance-penalized scoring metric is marginal in sparse samples, it is essential for optimal performance at convergence. 

The most important conclusion from the experiments is that when available, the guidance along with the score leads to a {\bf jump-start, better slope and in some cases, asymptotically sample efficient with a fraction of the number of instances needed} than merely learning from data. This clearly demonstrates the need for the injection of human guidance as knowledge when learning complex concepts.

\section{Conclusions}
We developed a human-in-the-loop one-shot concept learning framework in which the agent learns a generalized representation of a concept as FOL rules, from a single (or a few) positive example(s). Our 2 primary contributions are -- a new distance measure between concepts and richer human inputs than mere labels to be solicited actively by the agent. Our theoretical and experimental analyses showed the promise of \textsc{Goci} method. An exhaustive evaluation involving richer human inputs and integration with a plan induction system remain interesting directions of future research.



\bibliography{biblio}

\begin{thebibliography}{}

\bibitem[\protect\citeauthoryear{Bart and Ullman}{2005}]{bart2005single}
Bart, E., and Ullman, S.
\newblock 2005.
\newblock Single-example learning of novel classes using representation by
  similarity.
\newblock In {\em BMVC}.

\bibitem[\protect\citeauthoryear{Braziunas and
  Boutilier}{2006}]{BoutilierEtAl06}
Braziunas, D., and Boutilier, C.
\newblock 2006.
\newblock Preference elicitation and generalized additive utility.
\newblock In {\em AAAI}.

\bibitem[\protect\citeauthoryear{Chick}{2007}]{chick2007teaching}
Chick, H.~L.
\newblock 2007.
\newblock Teaching and learning by example.
\newblock {\em Mathematics: Essential research, essential practice}.

\bibitem[\protect\citeauthoryear{Das \bgroup et al\mbox.\egroup
  }{2018}]{DasEtAl18}
Das, M.; Odom, P.; Islam, M.~R.; Doppa, J.~R.; Roth, D.; and Natarajan, S.
\newblock 2018.
\newblock Preference-{G}uided {P}lanning: {A}n {A}ctive {E}licitation
  {A}pproach.
\newblock In {\em AAMAS}.

\bibitem[\protect\citeauthoryear{Degtyarenko \bgroup et al\mbox.\egroup
  }{2007}]{degtyarenko2007chebi}
Degtyarenko, K.; De~Matos, P.; Ennis, M.; Hastings, J.; Zbinden, M.; McNaught,
  A.; Alc{\'a}ntara, R.; Darsow, M.; Guedj, M.; and Ashburner, M.
\newblock 2007.
\newblock Chebi: a database and ontology for chemical entities of biological
  interest.
\newblock {\em Nucleic acids research} 36(suppl\_1):D344--D350.

\bibitem[\protect\citeauthoryear{Friend \bgroup et al\mbox.\egroup
  }{2018}]{pittphilsci14849}
Friend, M.; Khaled, M.; Lefever, K.; and Sz{\'e}kely, G.
\newblock 2018.
\newblock Distances between formal theories.

\bibitem[\protect\citeauthoryear{Fung, Mangasarian, and
  Shavlik}{2003}]{fung2003knowledge}
Fung, G.~M.; Mangasarian, O.~L.; and Shavlik, J.~W.
\newblock 2003.
\newblock Knowledge-based support vector machine classifiers.
\newblock In {\em NIPS}.

\bibitem[\protect\citeauthoryear{Goldman and
  Kuter}{2015}]{goldman2015measuring}
Goldman, R.~P., and Kuter, U.
\newblock 2015.
\newblock Measuring plan diversity: Pathologies in existing approaches and a
  new plan distance metric.
\newblock In {\em AAAI}.

\bibitem[\protect\citeauthoryear{Horn}{1951}]{horn1951sentences}
Horn, A.
\newblock 1951.
\newblock On sentences which are true of direct unions of algebras.
\newblock {\em The Journal of Symbolic Logic}.

\bibitem[\protect\citeauthoryear{Khan and Madden}{2014}]{khan2014}
Khan, S.~S., and Madden, M.~G.
\newblock 2014.
\newblock One-class classification: taxonomy of study and review of techniques.
\newblock {\em The Knowledge Engineering Review}.

\bibitem[\protect\citeauthoryear{Kozerawski and Turk}{2018}]{Kozerawski18CVPR}
Kozerawski, J., and Turk, M.
\newblock 2018.
\newblock Clear: Cumulative learning for one-shot one-class image recognition.
\newblock In {\em CVPR}.

\bibitem[\protect\citeauthoryear{Lake \bgroup et al\mbox.\egroup
  }{2011}]{lake2011}
Lake, B.; Salakhutdinov, R.; Gross, J.; and Tenenbaum, J.
\newblock 2011.
\newblock One shot learning of simple visual concepts.
\newblock In {\em CSS}.

\bibitem[\protect\citeauthoryear{Lake, Salakhutdinov, and
  Tenenbaum}{2015}]{lake2015human}
Lake, B.~M.; Salakhutdinov, R.; and Tenenbaum, J.~B.
\newblock 2015.
\newblock Human-level concept learning through probabilistic program induction.
\newblock {\em Science}.

\bibitem[\protect\citeauthoryear{Lisi}{2008}]{Lisi2008}
Lisi, F.~A.
\newblock 2008.
\newblock Building rules on top of ontologies for the semantic web with
  inductive logic programming.
\newblock {\em Theory and Practice of Logic Programming}.

\bibitem[\protect\citeauthoryear{Mitchell}{1980}]{Mitchell80}
Mitchell, T.~M.
\newblock 1980.
\newblock {\em The need for biases in learning generalizations}.
\newblock CS Dept, Rutgers Univ.

\bibitem[\protect\citeauthoryear{Mitchell}{1997}]{mitchell1997machine}
Mitchell, T.
\newblock 1997.
\newblock {\em Machine Learning}.
\newblock {McGraw-Hill Higher Education}.

\bibitem[\protect\citeauthoryear{Mooney}{1994}]{mooney1994preliminary}
Mooney, R.~J.
\newblock 1994.
\newblock {A Preliminary PAC Analysis of Theory Revision}.
\newblock {\em {Computational Learning Theory and Natural Learning Systems}}.

\bibitem[\protect\citeauthoryear{Muggleton and
  De~Raedt}{1994}]{muggleton1994inductive}
Muggleton, S., and De~Raedt, L.
\newblock 1994.
\newblock Inductive logic programming: Theory and methods.
\newblock {\em JLP}.

\bibitem[\protect\citeauthoryear{Muggleton and
  Feng}{1990}]{muggleton1990efficient}
Muggleton, S., and Feng, C.
\newblock 1990.
\newblock Efficient induction of logic programs.
\newblock In {\em ALT}.

\bibitem[\protect\citeauthoryear{MUGGLETON}{1988}]{Muggleton1988}
MUGGLETON, S.
\newblock 1988.
\newblock Machine invention of first-order predicates by inverting resolution.
\newblock In {\em Proc. of 5th Machine Learning Conference}.

\bibitem[\protect\citeauthoryear{Muggleton}{1991}]{muggleton1991inductive}
Muggleton, S.
\newblock 1991.
\newblock Inductive logic programming.
\newblock {\em New generation computing}.

\bibitem[\protect\citeauthoryear{Muggleton}{1995}]{muggleton1995}
Muggleton, S.
\newblock 1995.
\newblock Inverse entailment and progol.
\newblock {\em New generation computing}.

\bibitem[\protect\citeauthoryear{Narayan-Chen, Jayannavar, and
  Hockenmaier}{2019}]{minedialog}
Narayan-Chen, A.; Jayannavar, P.; and Hockenmaier, J.
\newblock 2019.
\newblock Collaborative dialogue in {M}inecraft.
\newblock In {\em ACL}.

\bibitem[\protect\citeauthoryear{Natarajan \bgroup et al\mbox.\egroup
  }{2009}]{natarajanilp}
Natarajan, S.; Kunapuli, G.; O’Reilly, C.; Maclin, R.; Walker, T.; Page, D.;
  and Shavlik, J.
\newblock 2009.
\newblock {ILP for bootstrapped learning: A layered approach to automating the
  ILP setup problem}.
\newblock In {\em ILP}.

\bibitem[\protect\citeauthoryear{Nau \bgroup et al\mbox.\egroup
  }{2003}]{nau2003shop2}
Nau, D.~S.; Au, T.-C.; Ilghami, O.; Kuter, U.; Murdock, J.~W.; Wu, D.; and
  Yaman, F.
\newblock 2003.
\newblock Shop2: An htn planning system.
\newblock {\em JAIR}.

\bibitem[\protect\citeauthoryear{Odom \bgroup et al\mbox.\egroup
  }{2015}]{odomaaai15}
Odom, P.; Khot, T.; Porter, R.; and Natarajan, S.
\newblock 2015.
\newblock Knowledge-based probabilistic logic learning.
\newblock In {\em AAAI}.

\bibitem[\protect\citeauthoryear{Pazzani}{1992}]{FOCL}
Pazzani, M.~J.
\newblock 1992.
\newblock An information-based approach to integrating empirical and
  explanation-based learning.
\newblock {\em ILP}.

\bibitem[\protect\citeauthoryear{Quinlan}{1990}]{Quinlan1990}
Quinlan, J.~R.
\newblock 1990.
\newblock Learning logical definitions from relations.
\newblock {\em Machine learning}.

\bibitem[\protect\citeauthoryear{Rouveirol}{1990}]{Rouveirol1990}
Rouveirol, C.
\newblock 1990.
\newblock Saturation: Postponing choices when inverting resolution.
\newblock In {\em ECAI}.

\bibitem[\protect\citeauthoryear{Rouveirol}{1992}]{Rouveirol1992}
Rouveirol, C.
\newblock 1992.
\newblock Extensions of inversion of resolution applied to theory completion.
\newblock {\em ILP}.

\bibitem[\protect\citeauthoryear{Sammut and Banerji}{1986}]{Sammut1986}
Sammut, C., and Banerji, R.~B.
\newblock 1986.
\newblock Learning concepts by asking questions.
\newblock {\em Machine learning: An artificial intelligence approach}.

\bibitem[\protect\citeauthoryear{Shavlik and Towell}{1989}]{shavlik89ebnn}
Shavlik, J.~W., and Towell, G.~G.
\newblock 1989.
\newblock Combining explanation-based learning and artificial neural networks.
\newblock In {\em ICML}.

\bibitem[\protect\citeauthoryear{Srinivasan}{2007}]{srinivasan2007aleph}
Srinivasan, A.
\newblock 2007.
\newblock Aleph: A learning engine for proposing hypotheses.
\newblock {\em Software available at http://web2. comlab. ox. ac.
  uk/oucl/research/areas/machlearn/Aleph/aleph. pl}.

\bibitem[\protect\citeauthoryear{Stroulia and
  Goel}{1994}]{stroulia1994learning}
Stroulia, E., and Goel, A.~K.
\newblock 1994.
\newblock Learning problem-solving concepts by reflecting on problem solving.
\newblock In {\em ECML}.

\bibitem[\protect\citeauthoryear{Tax}{2001}]{tax2002one}
Tax, D. M.~J.
\newblock 2001.
\newblock {\em One-class classification: Concept learning in the absence of
  counter-examples.}
\newblock Ph.D. Dissertation, TU Delft.

\bibitem[\protect\citeauthoryear{Towell and
  Shavlik}{1994}]{towell1994knowledge}
Towell, G.~G., and Shavlik, J.~W.
\newblock 1994.
\newblock Knowledge-based artificial neural networks.
\newblock {\em Artificial intelligence}.

\bibitem[\protect\citeauthoryear{Valiant}{1984}]{valiant1984theory}
Valiant, L.~G.
\newblock 1984.
\newblock A theory of the learnable.
\newblock {\em Communications of the ACM}.

\bibitem[\protect\citeauthoryear{Vapnik}{1999}]{vapnik1999overview}
Vapnik, V.~N.
\newblock 1999.
\newblock An overview of statistical learning theory.
\newblock {\em {IEEE Transactions on Neural Networks}}.

\bibitem[\protect\citeauthoryear{Vinyals \bgroup et al\mbox.\egroup
  }{2016}]{vinyals2016matching}
Vinyals, O.; Blundell, C.; Lillicrap, T.; Wierstra, D.; et~al.
\newblock 2016.
\newblock Matching networks for one shot learning.
\newblock In {\em {NIPS}}.

\bibitem[\protect\citeauthoryear{Vit{\'a}nyi \bgroup et al\mbox.\egroup
  }{2009}]{vitanyi2009normalized}
Vit{\'a}nyi, P.~M.; Balbach, F.~J.; Cilibrasi, R.~L.; and Li, M.
\newblock 2009.
\newblock Normalized information distance.
\newblock In {\em Information theory and statistical learning}. Springer.
\newblock  45--82.

\bibitem[\protect\citeauthoryear{Vit{\'a}nyi}{2013}]{vitanyi2013conditional}
Vit{\'a}nyi, P.~M.
\newblock 2013.
\newblock Conditional kolmogorov complexity and universal probability.
\newblock {\em Theoretical Computer Science}.

\bibitem[\protect\citeauthoryear{Wang \bgroup et al\mbox.\egroup
  }{2018}]{wang2018low}
Wang, Y.-X.; Girshick, R.; Hebert, M.; and Hariharan, B.
\newblock 2018.
\newblock Low-shot learning from imaginary data.
\newblock In {\em {CVPR}}.

\bibitem[\protect\citeauthoryear{Wilson and
  Latombe}{1994}]{wilson1994geometric}
Wilson, R.~H., and Latombe, J.-C.
\newblock 1994.
\newblock Geometric reasoning about mechanical assembly.
\newblock {\em Artificial Intelligence}.

\end{thebibliography}
\bibliographystyle{aaai}

\end{document}